\documentclass[journal]{IEEEtran}

\usepackage{fancyhdr}
\usepackage{cite}
\usepackage{amsmath,amssymb,bm}
\usepackage{array}
\ifCLASSOPTIONcompsoc
  \usepackage[caption=false,font=normalsize,labelfont=sf,textfont=sf]{subfig}
\else
  \usepackage[caption=false,font=footnotesize]{subfig}
\fi
\usepackage{url}
\usepackage{algorithm}
\usepackage{balance}
\usepackage{color}
\usepackage{epsfig}
\usepackage{epstopdf}
\usepackage{enumitem}
\usepackage{float}
\usepackage{graphicx}
\usepackage{setspace}
\usepackage{wasysym}
\usepackage{galois}
\usepackage{multirow}

\setenumerate[1]{itemsep=0pt,partopsep=0pt,parsep=\parskip,topsep=0pt}

\newfloat{figtab}{htb}{f}
\makeatletter
  \newcommand\figcaption{\def\@captype{figure}\caption}
  \newcommand\tabcaption{\def\@captype{table}\caption}
\makeatother
\usepackage[colorlinks, linkcolor=black, anchorcolor=black, citecolor=black]{hyperref}

\begin{document}
	
\title{Graph-based Hypothesis Generation for Parallax-tolerant Image Stitching}
\author{Jing Chen, \thanks{J. Chen is with the Center for Combinatorics, Nankai University, Tianjin 300071, China. Email: chenjing@mail.nankai.edu.cn.}
	 Nan Li$^*$, \thanks{L. Nan is with the Center for Applied Mathematics, Tianjin University, Tianjin 300071, China. Email: nan@tju.edu.cn.}
	 Tianli Liao, \thanks{T. Liao is with the Center for Combinatorics, Nankai University, Tianjin 300071, China. Email: liaotianli@mail.nankai.edu.cn.}}


\maketitle

\begin{abstract}
The seam-driven approach has been proven fairly effective for parallax-tolerant image stitching,
whose strategy is to search for an invisible seam from finite representative hypotheses of local alignment. In this paper, we propose a graph-based hypothesis generation and a seam-guided local alignment for improving the effectiveness and the efficiency of the seam-driven approach.
The experiment
demonstrates the significant reduction of number of hypotheses and the improved quality of naturalness of final stitching results, comparing to the state-of-the-art method SEAGULL.
\end{abstract}


\IEEEpeerreviewmaketitle

\section{Introduction}
\begin{figure}[H]
\centering
		\includegraphics[width=0.48\textwidth]{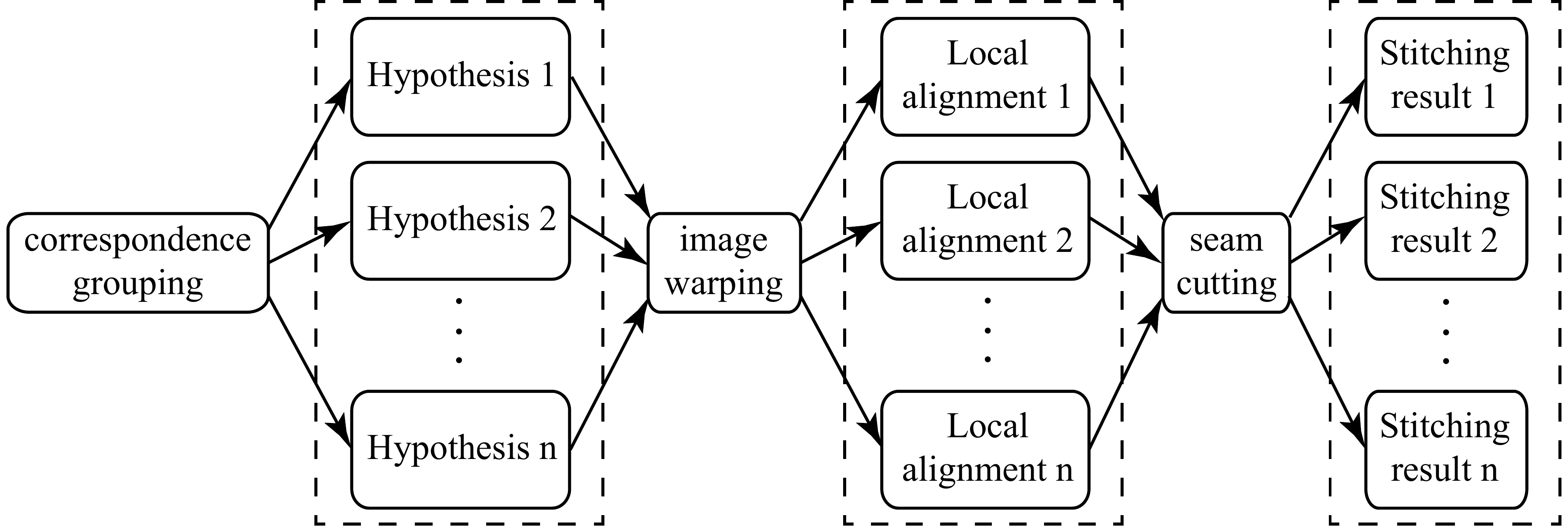}
  \caption{Overview of the seam-driven approach}
\label{fig:intro}
\end{figure}

Parallax handling is a challenging task for image stitching. Global alignment usually introduces noticeable artifacts or objectionable distortion. The seam-driven approach is a powerful tool for addressing the parallax problem, which searches finite representative hypotheses of local alignment for an invisible seam for plausible and natural stitching \cite{gao2013seam,zhang2014parallax,lin2016seam}. The effectiveness and the efficiency depend upon the quality and the number of hypotheses of local alignment, which correspond to image warping and correspondence grouping respectively (see Figure \ref{fig:intro}).

In fact, correspondence grouping includes to determine the warp and the strategy for aligning and grouping correspondences of feature. Image warping is evaluated by the qualities of seam, distortion and naturalness.
Gao \emph{et al.} \cite{gao2013seam} use homography (homo) and multiple RANSAC to divide the feature correspondences into hypotheses and generate corresponding local alignment via homography.
Zhang and Liu \cite{zhang2014parallax} use homography and a randomized strategy (seed) for correspondences grouping, then use the combination of homography and content-preserving warping (CPW) for generating hypotheses of local alignment. Lin \emph{et al.} \cite{lin2016seam} use homography and a new superpixel-based strategy (combinatorial) for correspondences grouping, then use the structure-preserving warping (SPW) with adaptive feature weighting for generating hypotheses of local alignment.
Because homography is not flexible enough, these correspondence grouping are not representative enough, then subsequently generate some undesirable hypotheses such that the efficiency becomes low.
On the other hand, these image warping suffer from distortion and naturalness issues in the cases of non-planar geometry of scenes,
such that their stitching results are less natural-looking.
Table \ref{table:intro} shows a comparison of seam-driven approaches in aspect of correspondence grouping and image warping.

\begin{table}[H]
	\centering
	\caption{Comparison of seam-driven approaches}
	\label{table:intro}
\vskip3pt
	\begin{tabular}{|c|c|c|}
		\hline
		Method & Correspondence grouping & Image warping  \\
		\hline
		Gao \emph{et al.} \cite{gao2013seam} &homo+multiple RANSAC & homo \\
        \hline
		Zhang \emph{\&} Liu \cite{zhang2014parallax} &homo+seed & homo+CPW \\
        \hline
		Lin \emph{et al.} \cite{lin2016seam} & homo+combinatorial & SPW \\
        \hline
		Ours & APAP+graph & SPMD \\
		\hline
	\end{tabular}
\end{table}
In this paper, we propose a graph-based hypothesis generation and a seam-guided local alignment for parallax-tolerant image stitching.
First, we use the as-projective-as-possible (APAP) warping and a graph-based strategy to group dual-feature correspondences into a few representative hypotheses.
Then we use a single-perspective mesh deformation (SPMD) with adaptive feature weighting that depends on current seams to generate corresponding local alignment.
 Finally, we search a good seam from these hypotheses and create the final stitching result.
Experiments
demonstrate significant reductions of number of hypotheses and improved qualities of naturalness of final stitching results,
comparing to SEAGULL. 

\section{Correspondence Grouping}

In order to increase the alignment accuracy and decrease the undesired distortion,
we consider both line segment and point correspondences as our feature correspondences,
which are called dual-feature in \cite{li2015dual}.
On the other hand, in order to increase the flexibility of warping,
we consider APAP \cite{zaragoza2013projective} as our grouping warp. In the following,
we formulate correspondence grouping into a weighted graph with source and sink,
then solve it via calculating the shortest path between them.
\begin{figure}[H]
        \includegraphics[width=0.48\textwidth]{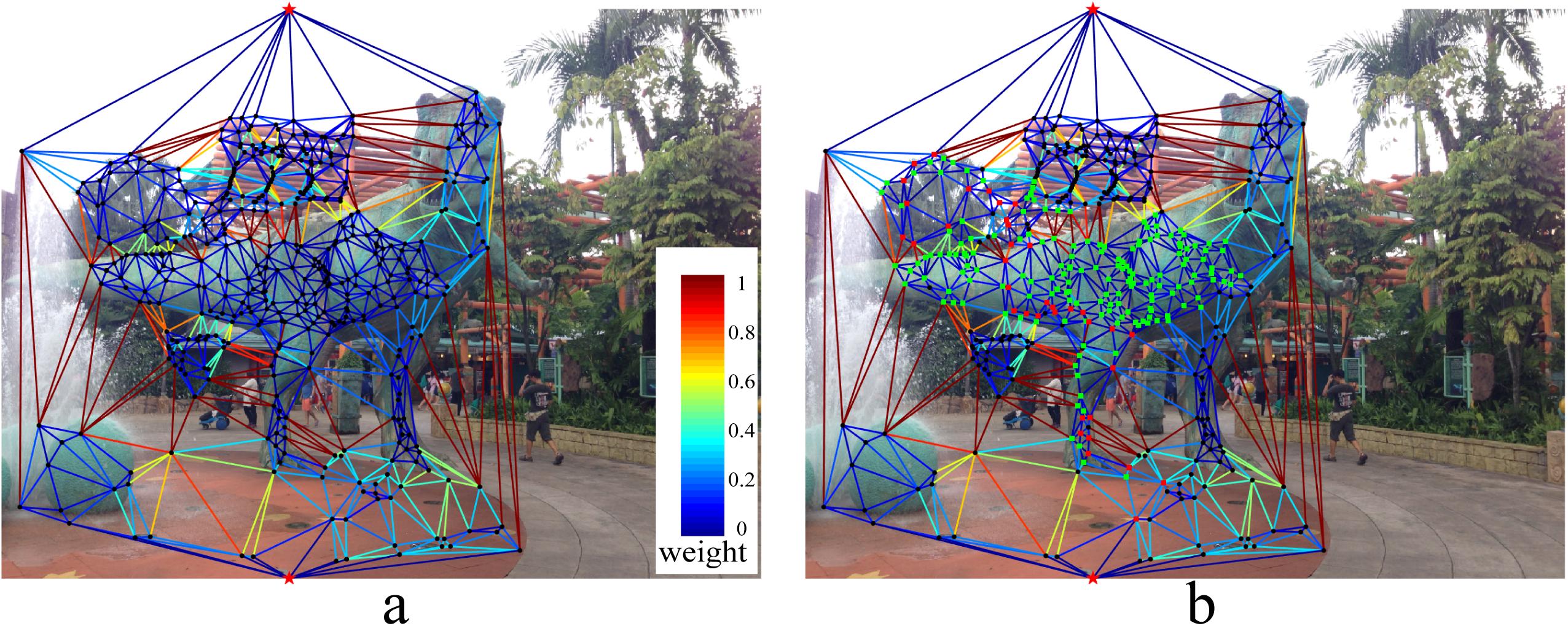}
  \caption{Overview of graph-based hypothesis generation. (a) Formulating (the edge weights are normalized into the range of $[0,1]$). (b) Grouping (the generated hypothesis consists of red and green vertices).}
\label{fig:con}
\end{figure}
For formulating, we first sample the line segment correspondence with three points in the reference image
(two endpoints and one midpoint) and insert a source point and a sink point on the image border
(up and down for horizontal stitching, or left and right for vertical stitching).
Then we set up a graph via Delaunay triangulation on the dual-feature points, the source and the sink points.
The weight for the edge between feature points $f_i$ and $f_j$ is defined by the cross residual distance error (CRDE),
which can be calculated by $\mathbf{w}$ (short for $\mathbf{w}(f_i,f_j)$),

\begin{equation}\label{apap_weight}
\mathbf{w}\!=\!\left\{
\begin{array}{cl}
  \|\mathbf{h}_{i}(f_{j})-f'_j\| + \|\mathbf{h}_{j}(f_{i})-f'_i\|, & f_i,f_j\in\mathrm{point}, \\
  d\left(\mathbf{h}_{i}(f_{j}),f'_j\right) + \|\mathbf{h}_{j}(f_{i})-f'_i\|, & f_i\in\mathrm{point},f_j\in\mathrm{line}, \\
  d\left(\mathbf{h}_{i}(f_{j}),f'_j\right) + d\left(\mathbf{h}_{j}(f_{i}),f'_i\right), & f_i,f_j\in\mathrm{line},\\
  0, & f_i\in\mbox{source or sink},
\end{array}
\right.
\end{equation}
where $\|\cdot\|$ is the Euclidean distance of points, $d(\,\cdot\,,\,\cdot\,)$ is the distance of line segments in \cite{li2015dual}
and $\mathbf{h}_i(f_j)$ is the transformed feature of $f_j$ under the location dependent homography $\mathbf{h}_i$ at the feature $f_i$,
which can be solved via the moving direct linear transformation (MDLT) approach by solving
\begin{equation}\label{mdltMatrix2}
   \mathrm{min}{\left\|\begin{bmatrix}{{\bf W}^{A}_{i}}~{{\bf W}^{B}_{i}}\end{bmatrix}{\begin{bmatrix}{\bf A}\\{\bf B}\end{bmatrix}{\bf h}_i}\right\|^2}~~~
   \text{s.t.}~~~\|{\bf h}_i\|=1,
 \end{equation}
where ${\bf A}\in\mathbb{R}^{2N\times9}$ is formed from point correspondences $\{(p_i,p'_i)\}_{i=1}^{N}$,
${\bf B}\in\mathbb{R}^{2M\times9}$ is formed from line segment correspondences $\{(l_j,l'_j)\}_{j=1}^{M}$,
${\bf W}^{A}_{i}\in\mathbb{R}^{2N\times2N}$ and ${\bf W}^{B}_{i}\in\mathbb{R}^{2M\times2M}$
depend on the distance between $f_i$ and other features $\{p_i\}_{i=1}^{N}$ and $\{l_j\}_{j=1}^{M}$.
Here, we refer to \cite{li2015dual,zaragoza2013projective,joo2015line} for more details.
Figure \ref{fig:con}(a) illustrates the weighted graph after formulating, where the edge weight somehow indicates
the flexibility to align the two feature correspondences via APAP. Therefore, we next need to determine a strategy to
group those correspondences into a hypothesis, which forms a subgraph that consists of the source and sink vertices with
edge weights less than a threshold.

For grouping, we first calculate the shortest path between the source vertex and the sink vertex in the weighted graph via Dijkstra's algorithm
(marked in red in Figure \ref{fig:con}(b)). Then we initialize the hypothesis by the features on the path
and iteratively insert new features that are not only connected to the last ones but also with edge weights less than a threshold.
The procedure terminates until no more new features can be inserted into the current hypothesis
(marked in green in Figure \ref{fig:con}(b)).
It is worth to note that, in order to avoid the shortest path from running into a path with less vertices
but larger weight than the threshold, we use a sigmoid metric to preprocess the edge weights,
which is similar to the metric in \cite{Li2018}.

Other hypotheses are subsequently generated by iteratively modifying the edge weights of the last shortest path to
a very large positive number, and performing the procedure of
grouping until the number of remaining correspondences is less than $30$,
or the edge weight of the shortest path is larger than the threshold.
In the experiment, we use VLFeat to extract and match SIFT features,
use LSD to extract line segments and match them by \cite{jia2016novel} and
use RANSAC to remove outliers.
The threshold of edge weights is set to $10$.
Because APAP is more flexible than homography, our graph-based correspondences grouping
generates less but more representative hypotheses.
A comparison of number of hypotheses with SEAGULL \cite{lin2016seam} is illustrated in Table \ref{table:exp}.

\section{Image Warping}
In order to address distortion and naturalness issues, we consider SPMD \cite{Li2018tmm} as our image warping to
generate corresponding local alignment for each hypothesis.
First, we mesh the target image by vertices indexing from $1$ up to $n$ and reshape them into a $2n$-dimension vector $V$.
The $n$ vertices after deformation are denoted by $\hat{V}$. Then the total energy function is defined by
\begin{equation}\label{eq_energy}
E(\hat{V})=\lambda_\mathrm{p} E_\mathrm{p}(\hat{V}) + \lambda_\mathrm{l} E_\mathrm{l}(\hat{V}) + E_{\mathrm{cl}}(\hat{V}) + \lambda_\mathrm{s} E_\mathrm{s}(\hat{V}),
\end{equation}
where these four terms
address the alignment, naturalness, distortion and saliency issue respectively.
Finally, local alignment is obtained by solving $\min E(\hat{V})$ via
any sparse linear solver, since (\ref{eq_energy}) is sparse and quadratic.
Here, we refer to \cite{Li2018tmm} for more details.

In order to further improve the quality of seam of local alignment, we use the adaptive feature weighting strategy in \cite{lin2016seam}
such that the process of optimizing local alignment is guided by the estimated seam.
Respectively, alignment and naturalness terms are modified to
\begin{equation}\label{eq_align}
E_\mathrm{p}(\hat{V})  = \sum_{i=1}^N w_i\|\varphi(\hat{p}_i)-p'_i\|^2,~E_\mathrm{l}(\hat{V})  = \sum_{j=1}^M w'_jd\left(\varphi(\hat{l}_j),l'_j\right)^2,
\end{equation}
where $\varphi(\cdot)$ is the bilinear interpolation of the four enclosing grid vertices of the dual-feature point,
$w_i$ and $w'_j$ are the adaptive feature weights that depend on current alignment errors and distances to current seams.
Here, we refer to \cite{lin2016seam} for more details.

For the estimation of the stitching seam, we use the perception-based seam-cutting approach in \cite{Li2018},
which uses a sigmoid metric to characterize the perception of color discrimination and
a saliency weight to simulate that the human eye inclines to pay more attention to the salient objects.

The iteration terminates until the average change of vertex locations compared to the last iteration
is less than one pixel or the iteration number exceeds $5$.
In the experiment, the parameters are set to the recommended values in \cite{lin2016seam,Li2018,Li2018tmm}.
Because characteristics of local alignment,
naturalness, distortion and
saliency are simultaneously and iteratively optimized,
our seam-guided local alignment
create more natural-looking final stitching results.
Comparisons of qualities of seam and naturalness of hypotheses with
SEAGULL
are illustrated in Table \ref{table:exp} and Figure \ref{fig:exp}.

\section{Experiments}
We compare our proposed method with the state-of-the-art method SEAGULL \cite{lin2016seam}. We evaluate
these two methods over the publicly available dataset of $24$ pairs of images with challenging parallax variation
from SEAGULL\footnote{\url{http://linkaimo.com/publications/ImageStitching/ImageStitching.html}}.

Firstly, we compare the number of hypotheses of our correspondence grouping with SEAGULL.
Table \ref{table:exp} shows the comparison results, where in most of the cases,
our grouping method generates the smaller number of hypotheses.
The comparison results for other seam-driven approaches \cite{gao2013seam,zhang2014parallax} over the same dataset
can be found in \cite{lin2016seam}.

\begin{table}[H]
\caption{Quantitative comparison between SEAGULL and our method. Bold values indicate best results}
\centering
	\label{table:exp}
	\renewcommand\arraystretch{1.3}
	\vskip3pt
\begin{tabular}{p{0.26cm}|p{0.35cm}p{0.5cm}|p{0.35cm}p{0.6cm}||p{0.26cm}|p{0.35cm}p{0.5cm}|p{0.35cm}p{0.5cm}}
		\hline
		\multirow{2}[3]{*}{No.} & \multicolumn{2}{c|}{SEAGULL} & \multicolumn{2}{c||}{Ours} & \multirow{2}[3]{*}{No.} & \multicolumn{2}{c|}{SEAGULL} & \multicolumn{2}{c}{Ours} \\
		\cline{2-5}\cline{7-10}          & hypo  & seam  & hypo  & seam  &       & hypo  & seam  & hypo  & seam \\
		\hline
		01.   &  3     &  0.148     & \textbf{1}      & \textbf{0.136}     & 13.   &  3     &  0.045     &  \textbf{2}     &  \textbf{0.010}\\
		02.   &  1     &  0.061     & 1      & \textbf{0.043 }    & 14.   &  \textbf{1}     &  0.074     &  2     &  \textbf{0.069} \\
		03.   &  2    &  0.135     & 2      & \textbf{0.127}      & 15.   &    5   &  \textbf{0.205}     & \textbf{3}      &  0.265\\
		04.   &  3     &  0.217     & \textbf{2}      & \textbf{0.186}      & 16.   & 3      &   0.138   &    \textbf{1}   & \textbf{0.117} \\
		05.   &  \textbf{1}     &  0.387     & 2      & \textbf{0.342}      & 17.   &  10     &\textbf{0.114}   &   \textbf{3}    & 0.185 \\
		06.   &  6     &  0.072     & \textbf{2}      & \textbf{0.068}   & 18.   &    7  &     0.336  & \textbf{2}    & \textbf{0.287} \\
		07.   &  6     &  0.168     & \textbf{2}      &\textbf{0.159}      & 19.   &  3     &   0.142    & \textbf{2}      &  \textbf{0.126}\\
		08.   &  5     & 0.072     & \textbf{3}     &  \textbf{0.065}    & 20.   & 3      & 0.170      &    \textbf{2}  &  \textbf{0.167}\\
		09.   &  5     &  0.066     &  \textbf{2}     & \textbf{0.062}    & 21.   & 2      &   0.179    &      2 & \textbf{0.171}  \\
		10.   &  7     &  \textbf{0.195}     & \textbf{3}      & 0.226    & 22.   &    9   &   0.080    & \textbf{2} &  \textbf{0.076} \\
		11.   &  1     &  0.256     & 1      &\textbf{0.239}    & 23.   & 13      &   0.159    &  \textbf{4}     & \textbf{0.130} \\
		12.   &  6     &  0.265     & \textbf{3}      &\textbf{0.237}     & 24.   & 6      &  0.148     &    \textbf{2}   &  \textbf{0.136} \\
		\hline
	\end{tabular}%
\end{table}

\begin{figure}[H]
	\centering
    \includegraphics[width=0.48\textwidth]{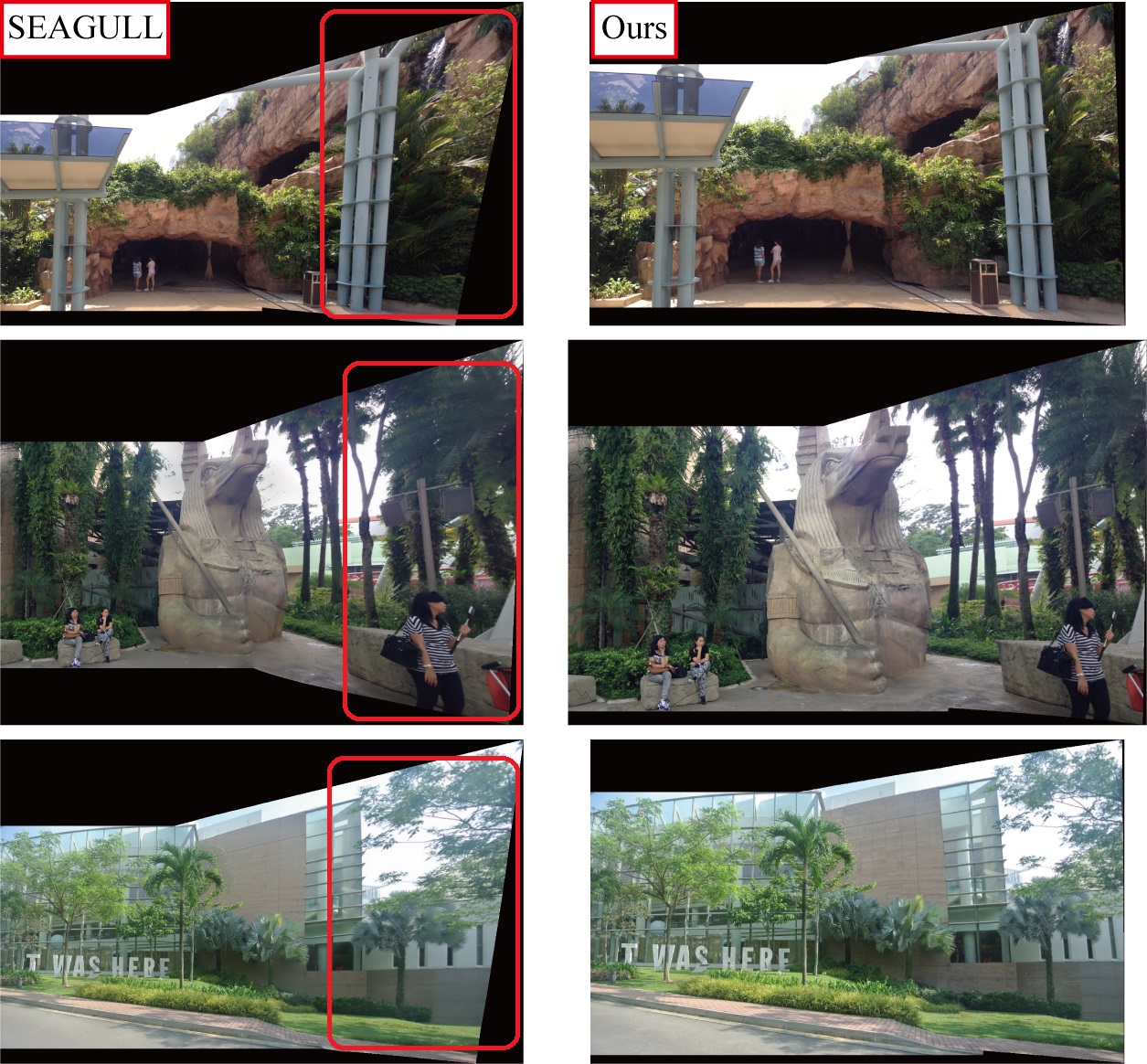}
	\caption{Qualitative comparison between SEAGULL and our method.}
\label{fig:exp}
\end{figure}

We also compare the seam and the naturalness qualities of our image warping with SEAGULL.
Table \ref{table:exp} including the number of hypotheses (hypo) and the seam quality (seam) shows the comparison results of seam quality,
which is measured by the zero-mean normalized cross correlation (ZNCC) score of $15\times15$ local patches along the seam as in \cite{lin2016seam}. In most of the cases,
our warping method has the lower score of local alignment.
Figure \ref{fig:exp} illustrates the comparison results of final stitching results, where SEAGULL suffers from
projective distortion (indicated in red rectangles) at times while our final stitching results are more natural-looking.
All $24$ pairs of comparison results are available in the supplementary material.

%

\section{Conclusions}

In this paper, we proposed a novel graph-based hypothesis generation and a seam-guided local alignment for parallax-tolerant image stitching.
Experiments
demonstrated significant reductions of number of hypotheses and improved qualities of naturalness of final stitching results,
comparing to the state-of-the-art method SEAGULL.

\vskip6pt




\bibliographystyle{IEEEtran}

\vskip6pt

\vskip3pt
\begin{thebibliography}{}
\bibitem{gao2013seam}
Gao J., Li Y., Chin T.-J., and Brown M. S.: `Seam-driven image stitching',
\textit{Eurographics}, 2013, pp. 45--48

\bibitem{zhang2014parallax}
Zhang F., and Liu F.: `Parallax-tolerant image stitching', \textit{Proc. IEEE
  Conf. Comput. Vision Pattern Recognit.}, 2014, pp. 3262--3269

\bibitem{lin2016seam}
Lin K., Jiang N., Cheong L.-F., Do M., and Lu J.: `{SEAGULL}: Seam-guided
  local alignment for parallax-tolerant image stitching', \textit{Proc. Eur.
  Conf. Comput. Vis.}, 2016, pp. 370--385

\bibitem{li2015dual}
Li S., Yuan L., Sun J., and Quan L.: `Dual-feature warping-based motion model
  estimation', \textit{Proc. IEEE Int. Conf.
  Comput. Vis.}, 2015, pp. 4283--4291

\bibitem{zaragoza2013projective}
Zaragoza J., Chin T.-J., Brown M. S., and Suter D.: `As-projective-as-possible
  image stitching with moving {DLT}', \textit{IEEE Trans.
Pattern Anal. Mach. Intell.}, 2014, {\bf 7}(36), pp. 1285--1298

\bibitem{joo2015line}
Joo K., Kim N., Oh T. H., and Kweon I. S.: `Line meets as-projective-as-possible image stitching with moving {DLT}', \textit{Proc. IEEE Int. Conf. Image Process.}, 2015, pp. 1175--1179

\bibitem{Li2018}
Li N., Liao T., and Wang C.: `Perception-based seam cutting for image stitching',
\textit{SIViP},
2018, doi:10.1007/s11760-018-1241-9


\bibitem{jia2016novel}
Jia Q., Gao X., Fan X., Luo Z., Li H., and Chen Z.: `Novel coplanar
line-points invariants for robust line matching across views', \textit{Proc.
Eur. Conf. Comput. Vis.}, 2016, pp. 599--611

\bibitem{Li2018tmm}
Li N., and Liao T.: `Single-perspective warps in natural image stitching',
arXiv:1802.04645,
2018

\end{thebibliography}

\end{document}